\documentclass[11pt,letterpaper]{article}
\usepackage{emnlp2017}
\usepackage [english]{babel}
\usepackage [autostyle, english = american]{csquotes}
\MakeOuterQuote{"}
\usepackage{times}
\usepackage{latexsym}
\usepackage{color}
\usepackage{colortbl}
\usepackage[colorlinks]{}
\usepackage{pgfplots,wrapfig}
\usepackage{tikz}
\usepackage{booktabs}
\newcommand{\ra}[1]{\renewcommand{\arraystretch}{#1}}
\newcolumntype{P}[1]{>{\centering\arraybackslash}p{#1}}
\pgfplotsset{compat=1.12}
\usepackage{array, caption, floatrow, tabularx, makecell, booktabs}%
\usepackage{hyperref}

\usepackage{subcaption}

\usepackage{url}

\emnlpfinalcopy

\def\emnlppaperid{***}

\title{Harvesting Creative Templates \\ for Generating Stylistically Varied Restaurant Reviews}

\author{Shereen Oraby, Sheideh Homayon, \and Marilyn Walker \\
Natural Language and Dialogue Systems Lab \\
University of California, Santa Cruz \\
  {\tt \{soraby,shomayon,mawalker\}@ucsc.edu}}

\date{}

\begin{document}

\maketitle

\begin{abstract}
Many of the creative and figurative elements that make language
exciting are lost in translation in current natural language
generation engines. In this paper, we explore a method to harvest
templates from positive and negative reviews in the restaurant domain,
with the goal of vastly expanding the types of stylistic variation
available to the natural language generator. We learn hyperbolic
adjective patterns that are representative of the strongly-valenced
expressive language commonly used in either positive or negative
reviews.  We then identify and delexicalize entities, and use
heuristics to extract generation templates from review sentences. We
evaluate the learned templates against more traditional review
templates, using subjective measures of {\it convincingness}, {\it
  interestingness}, and {\it naturalness}. Our results show that the
learned templates score highly on these measures.  Finally, we analyze
the linguistic categories that characterize the learned positive and
negative templates. We plan to use the learned templates to improve the
conversational style of dialogue systems in the
restaurant domain.
\end{abstract}

\section{Introduction}

\begin{table}[t!h]
\begin{footnotesize}
\begin{tabular}
{@{}p{0.1cm}|p{0.5cm}|p{6.1cm}@{}}
\toprule
{\bf \#} & {\bf Stars} & {\bf Review} \\ \midrule
1 & {\sc 1/5} & This place is probably the worst thing that ever happened to the history of the known world. [...]  The food, however, I initially would want to call unremarkable but I can't.  I can't call it unremarkable because it is so incredibly remarkably terrible. [...] \\ \midrule
2 & {\sc 2/5} & Can't say anything about the food, as we were never served.  We never saw a server, even after sitting at our table for 15 minutes. Unacceptable. \\ \midrule
3 & {\sc 3/5} & I was back here a couple of days ago with my family. And although I remember The food being a lot better than this time around. I was kind of disappointed. The service was okay since I had no Jose this time. Nothing to mention here just refills chips salsa and beverages when you need and food when it's ready. \\ \midrule
4 & {\sc 4/5} & I would eat here everyday if I didn't think I'd end up 400 pounds... Minus 1 star because each time I've been here the service has kinda sucked and orders have been messed up. Regardless, their fried chicken on waffles topped with syrup and a slice of Red Velvet cake to top it off......... is sooooooo heavenly. \\ \midrule
5 & {\sc 5/5} & I only have one warning about this restaurant.  The food is so amazing that you cannot eat Mexican food anywhere else.  [...] I had chicken and beef enchiladas which had homemade corn tortillas and the most tender meat I had ever tasted.  [...] I will be a customer for life here! \\ 
\bottomrule         
 \end{tabular}
 \caption{Restaurant Reviews by Rating from the Yelp Dataset Challenge Corpus}
   \label{table:yelp}
 \end{footnotesize}
\end{table}

The restaurant domain has been one of the most common applications for
spoken dialogue systems for at least 25 years
\cite{PHSZ92,WWM02,StentPrasadWalker04,devillers2004french,gavsic2008training}.
There has been a tremendous amount of previous work on natural
language generation of recommendations and descriptions for
restaurants \cite{Howcroftetal13,Wenetal15,novikova2016crowd}, some of
which has even focused on generating stylistically varied restaurant
recommendations
\cite{Higashinakaetal07,MairesseWalker10,dethlefs2014cluster}. Given
this, it is surprising that previous work has not especially noted
that restaurant reviews are a fertile source of creative and
figurative language. For example, consider the elaborate descriptions
in the restaurant reviews in Table \ref{table:yelp}\footnote{Reviews
  from the Yelp 2016 dataset challenge: \\ \url{https://www.yelp.com/dataset_challenge}}, e.g. phrases such as {\it worst thing that ever happened in the history of the known world} along with {\it incredibly remarkably terrible} (Row 1),  {\it eat here everyday if I didn't think I'd end up 400 pounds} and {\it sooooooo heavenly} (Row 4), and {\it
  food so amazing you cannot eat [...] anywhere else} (Row 5).
These phrases express
extremely valenced reactions to restaurants, their menu items, and
related attributes, using figurative language.

\begin{table*}[ht]
\begin{footnotesize}
\begin{tabular}
{@{}p{3.2cm}|p{7.8cm}|p{3.8cm} @{}}
\toprule
{\bf  MR } & {\bf NLG } & \bf System \\ \midrule
{ 
name[Emilios]\newline
foodquality[excellent]\newline
location[city centre]\newline
cuisine[Italian]\newline
decor[decent]\newline
service[decent]}
 & {\it EmilioÕs decor and service are both decent, but its food quality is nothing short of excellent. It serves Italian food and its in the City Centre.} & Seq2Seq NLG \newline \cite{Nayaketal17} \\ \hline
{ name[The Eagle]\newline
eattype[coffee shop]\newline
pricerange[moderate]\newline
customerrating[3/5]\newline
kidsFriendly[yes]\newline
near[Burger King]}  & 
{\it The three star coffee shop, The Eagle, gives families a mid-priced dining experience featuring a variety of wines and cheeses. Find The Eagle near Burger King.} & Human annotators \newline \cite{novikova2016crowd} \\ \hline
{ name[X]\newline
type[placetoeat]\newline
near[Y]\newline
food[Continental]\newline
food[French]}
&  {\it X is a restaurant providing french and continental food, near Y.} & Seq2Seq Generation \newline \cite{Dusek2016}  \\ \hline
{ name[Red door cafe]\newline
goodformeal[breakfast]\newline
area[cathedral hill]\newline
kidsallowed[no]}
&  {\it Red door cafe is a good restaurant for breakfast in the area of cathedral hill and does not allow children . }& LSTM Generation \newline \cite{Wenetal15}  \\ \hline
{ name[Le Marais]\newline 
cuisine[French]\newline
foodquality[good]\newline
service[good]\newline
decor[decent]\newline
price[\$44]} &  {\it I am sure you would like Le Marais, you know. The atmosphere is acceptable, the servers are nice and it's a fresh, kosher and steak house place. Actually, the food is good, even if its price is 44 dollars.} & {\sc Personage} \newline \cite{Mairesse07} \\ \hline
{ name[Babbo]\newline
foodquality[superb]\newline
service[excellent]\newline
decor[superb]} &  {\it The food is phenomenal and the atmosphere is very unique. Babbo has excellent service. It has the best overall quality among the selected restaurants.} & Unsupervised Method for \newline Lexicon Learning \newline \cite{Higashinakaetal06} \\ 
\bottomrule         
 \end{tabular}
 \caption{Example Meaning Representations (MR) and
Corresponding Natural Language Generation (NLG) Output in the Restaurant Domain}
   \label{table:MR}
 \end{footnotesize}
\end{table*}

The creativity exhibited in these user-generated restaurant reviews 
can be contrasted with  natural language
generation (NLG) for the restaurant domain.  Methods for NLG
typically begin with a structured meaning representation
(MR), as shown in Table~\ref{table:MR}, and map these meaning
representations into surface language forms, using a range of
different methods, including template-based generation, statistically
trained linguistically-informed NLG engines, and neural approaches \cite{BangaloreRambow00,WalkerRambow02}. 
These approaches vary in the degree to which they can generate
syntactically and semantically correct utterances, but in most cases
the stylistic variation they can generate is extremely limited.
Table~\ref{table:MR} illustrates sample restaurant domain  utterances produced by 
recent statistical/neural natural language generators \cite{Higashinakaetal06,Mairesse07,Wenetal15,novikova2016crowd,Dusek2016}.

One of the most prominent characteristics of restaurant reviews in the
Yelp corpus is the prevalent use of hyperbolic language, such as the
phrase {\it "incredibly remarkably terrible"} in Table~\ref{table:yelp}. Hyperbole
is often found in persuasive language, and is classified 
as a form of figurative language
\cite{McCarthy2004,CanoMora2009}. Colston and O'Brien describe how an event or
situation evokes a scale, and how hyperbole exaggerates a literal situation, 
introducing a discrepancy
between the "truth" and what is said \cite{ColstonKeller98,ColstonObrien00b}.  
Hyperbole moves the
strength of a statement up and down the scale, away from the literal
meaning, where the degree of movement reflects the degree of contrast
or exaggeration. Depending on what they modify, adverbial intensifiers
like {\it totally, absolutely,} and {\it incredibly} can shift the strength of the
assertion to extreme negative or positive. 

Similarly, \newcite{KreuzRoberts95} describe a standard frame for hyperbole in
English where an adverb modifies an extreme, positive adjective,
e.g. {\it "That was {\bf absolutely amazing}!"} or {\it "That was {\bf
    simply the most incredible} dining experience in my entire life."}
Such frames can be seen in the reviews in Table~\ref{table:yelp}, but
we also see many other idiomatic hyperbolic expressions such as {\it
  out of this world} \cite{CanoMora2009}.

Our goal is to develop a natural language generator for the restaurant
domain that can harvest and make use of these types of stylistic
variations.  We explore a data-driven approach to automatically select stylistically
varied utterances in the restaurant review domain as candidates for
review construction. We empirically learn hyperbolic adjective
patterns that are highly correlated with two classes (positive and
negative reviews). Using different resources, we also identify and
delexicalize {\it restaurant, cuisine, food, service,} and {\it staff}
entities, and select short, single-entity utterances that are simple
to templatize.

Our overall approach is thus similar to
\newcite{Higashinakaetal06,Higashinakaetal07}, who describe a method
for harvesting an NLG dictionary from restaurant reviews, however our
focus on learning expressive language, in particular hyperbole as a
type of figurative language, is novel.  Our framework consists
of the following steps:

\begin{enumerate}\itemsep0em
\item Collect a large number of strongly positive and strongly negative reviews in the restaurant domain;
\item Use a linguistic pattern learner to identify linguistic frames that use hyperbole;
\item Create generation templates from  the identified linguistic patterns and infer their contexts of use;
\item Learn to rank the generation templates for convincingness and
  quality.
\end{enumerate}

We see Steps 1 to 3 as the overgeneration phase, aimed at
vastly expanding the types of stylistic variation possible, while Step 4 is
the ranking phase, in a classic overgenerate and rank NLG architecture
\cite{LangkildeKnight98,RambowRogatiWalker01}.  We focus in this paper
on Steps 1 to 3, expecting to improve these steps before we
move on to Step 4. 

Thus, in this paper, we conducted an evaluation experiment to 
compare three different types of NLG templates: pre-defined {\sc
  basic} templates similar to those used in current NLG engines for
the restaurant domain \cite{Walkeretal07,Wenetal15}, the basic
templates stylized with our learned patterns for more {\sc hyperbolic}
templates, and finally a class of {\sc creative} templates that
incorporate full sentence templates from user reviews. Our
expectation was that many of the {\sc creative} templates would fail
to be appropriate to their contexts, but that our {\sc hyperbolic}
templates would be both appropriate and more interesting and
convincing than the {\sc basic} templates.  However, our results show
that our creative templates are preferred as more convincing,
interesting, and natural across the board. We discuss how we
can use quantitative metrics associated with the learned templates for future ranking, and analyze characteristic
 linguistic categories in each class.

\section{Data}

Our restaurant review data comes from the Yelp dataset challenge,
which includes 144K businesses with over 4.1M reviews. We randomly
select 10K businesses located in the US that are classified as
restaurants, resulting in a set of around 40K reviews. The data
consists of around 4K 1 stars, 3.8K 2 stars, 5.6K 3 stars, 11.3K 4
stars, and 15K 5 stars. We divide the reviews by stars, and create
three datasets: negative (using all of the 1-2 stars), positive
(balancing the number of negative reviews using the 5 stars), and
neutral (using all of the 3 stars). Table \ref{table:data} shows our
data distribution.

\begin{table}[h]
\begin{footnotesize}
\begin{tabular}
{@{}P{2cm}|P{1cm}|P{2.5cm}@{}}
\toprule
\bf Split & \bf Stars & \bf Num Reviews \\ \midrule
{\sc Positive} & 5 & 7,853 \\
{\sc Neutral}& 3 & 5,610 \\ 
{\sc Negative}& 1-2 & 7,853 \\          
\bottomrule
 \end{tabular}
 \caption{Selected Review Data Distribution}
   \label{table:data}
 \end{footnotesize}
\end{table}

\section{Learning Patterns for Hyperbole}
Our goal is to learn patterns that are highly associated with the
extreme positive and negative reviews, and that exemplify strong,
expressive language. To automatically learn such patterns, we use the
AutoSlog-TS weakly-supervised extraction pattern learner
\cite{Riloff96}. 

AutoSlog-TS uses a set of syntactic templates to learn
lexically-grounded patterns.
AutoSlog does not require fine-grained labels on training data: all it
requires is that the training data be divided into two distinct
classes. Here, we run two separate AutoSlog experiments, one in which
the classes are {\sc positive} compared to {\sc neutral}, and the
other where the {\sc negative} class is compared to {\sc neutral}.  We
hypothesize that in this way, we can surface the most commonly used
patterns from each class that are not necessarily sentiment-related. 

AutoSlog applies the Sundance shallow parser \cite{RiloffPhillips04}
to each sentence of each review, finds all possible matches for its
syntactic templates, and then instantiates the syntactic templates with
the words in the sentence to produce a specific lexico-syntactic
expression. Most importantly, it uses the labels associated with the
data to compute statistics for how frequently each pattern occurs in
each class. Thus, for each pattern $p$, we learn the P({\sc
  positive/negative}$\mid p$), the P({\sc neutral}$\mid p$), and
the pattern's frequency.

 Table \ref{table:aslog} shows examples of the patterns we learn and
 sample instantiations, with their respective frequency (F) and
 probabilities (P). In the pattern template column of
 Table~\ref{table:aslog}, PassVP refers to passive voice verb phrases
 (VPs), ActVP refers to active voice VPs, InfVP refers to infinitive
 VPs, and AuxVP refers to VPs where the main verb is a form of {\it to
   be} or {\it to have}. Subjects (subj), direct objects (dobj), noun
 phrases (np), and possessives (genitives) can also be extracted by
 the patterns.  Because we are particularly interested in descriptive
 patterns, we also use ngram pattern templates, {\tt AdjAdj,
   AdvAdj, AdvAdvAdj}, as in related work
 \cite{Orabyetal15,Orabyetal16}.

\begin{table}[t]
\begin{scriptsize}
\begin{tabular}
{@{}p{0.3cm}|p{0.2cm}|P{2.8cm}|P{2.8cm}@{}}
\toprule
 \bf F & \bf P &  {\sc {\bf Pattern Template}} & {\sc {\bf Example Pattern}} \\
    \hline
    \multicolumn{3}{l}{ \bf Positive} \\ \midrule
  40    & 1.0 &     $<$subj$>$ActInfVP Dobj & $<$subj$>$ wait come \\
  19 & 1.0 & ActVP Prep $<$Np$>$ &  tucked in $<$Np$>$ \\
  54    & 0.9 & AdjAdj & hands down \\
  30 & 0.9 & $<$subj$>$ActVP Dobj & $<$subj$>$ worth wait \\
  20 & 0.9 & NpPrep $<$Np$>$ & screaming for \\
    10  & 0.9 &     $<$subj$>$ AuxVP Adj & $<$subj$>$ be scrumptious \\ 
    416 & 0.8 & AdjNoun & great food \\
 16  & 0.8 &    PassVP Prep NP &  addicted to \\
 113 & 0.7 & AdvAdj & very fresh \\
 4 & 0.7 & AdjNoun & go-to restaurant \\
\midrule
    \multicolumn{3}{l}{ \bf Negative} \\ \midrule
    17 & 1.0 & $<$subj$>$ AuxVP Adj & $<$subj$>$ be impossible \\
  13    & 1.0 & AdjNoun & negative stars \\
  12    & 1.0 & $<$subj$>$ ActVP Dobj & $<$subj$>$ got poisoning \\
  23    & 0.9 & AdjNoun & no sense \\
  134 & 0.8 & $<$subj$>$ AuxVP Adj & $<$subj$>$ be awful \\
  26 & 0.8 & $<$subj$>$ AuxVP Adj & $<$subj$>$ be rubbery \\
19   & 0.8 &    $<$subj$>$ ActVP   & $<$subj$>$ not waste \\
107 & 0.8 & AdjNoun & poor service \\
100 & 0.8 & AdjNoun & no way \\
201 & 0.8 & $<$subj$>$ AuxVP Adj & $<$subj$>$ not be back \\
\bottomrule
 \end{tabular}
 \caption{Examples of Pattern Templates in AutoSlog-TS and Instantiations by Class}
   \label{table:aslog}
 \end{scriptsize}
\end{table}

Our goal is to find highly reliable patterns without sacrificing
linguistic variation. Current statistical methods for training NLG
engines typically eliminate linguistic variability by seeking to learn
standard, more generic patterns that occur frequently in the data
\cite{Liuetal16b,Nayaketal17}.
 Since this phase of our work aims to
vastly expand  the amount of linguistic variation possible, we select
instantiations that have a frequency of at least 3, and a probability
of at least 0.75 association with the respective class \cite{Orabyetal15,Orabyetal16}. 
We hypothesize that patterns that
occur at least 3 times should be fairly reliable, and those that have
at least a 75\% probability of being associated with the positive or
negative class should be distinctive.  Using these filters, we learn
8,320 positive adjective patterns, and 7,839 negative adjective
patterns. 

We also observe that patterns learned
using stricter thresholds (for example, frequency of at least 10 and probability
of at least 0.9) also gives us useful patterns, and note that we can use the
frequencies and probabilities in our future rank task. For larger coverage, we
experiment with our less restrictive thresholds in the current work.

\section{Designing Review Templates}

To make use of the descriptive adjective patterns we learned, we needed to first identify what
entities each of the patterns describes. To do this, we aggregate
lexicons for each of five important restaurant entities: {\it
  restaurant-type, cuisine, food, service,} and {\it staff} using
Wikipedia\footnote{\url{https://www.wikipedia.org/}} and
DBpedia\footnote{\url{http://wiki.dbpedia.org/}}. We end up with 14
items for {\it restaurant-types} (e.g. "cafe"), 45 for {\it cuisines}
(e.g. "Italian"), 4,913 for {\it foods and ingredients}
(e.g. "sushi"), 12 for {\it staff} (e.g. "waiters"), and 2 for {\it
  service} (e.g. "customer service").

\subsection{Basic Templates}
To construct the most basic set of templates, we use simple relationships between adjectives and the entities they describe to define a set of sentences with entity slots, i.e. {\it "They had [adj] (entity)."}, {\it "The (entity) was$\mid$is [adj]."}, {\it "The (entity) looked$\mid$tasted [adj]."} We use basic lists of adjectives commonly found in reviews for these baseline templates. To vary the templates, we alternate between using only simple sentences, and sometimes combine related entities into more complex sentences (e.g. {\it service} and {\it staff}, or {\it restaurant-type} and {\it cuisine}).

\subsection{Hyperbolic Templates}

For our hyperbolic templates, we replace the standard adjectives in
the basic templates with  adjective patterns learned from
the  restaurant reviews. To select appropriate adjectives patterns for replacement
in each basic template, we first delexicalize the sentences that
instantiate our learned adjective patterns for each class, and create
sets of (entity, adjective pattern) pairs based on the relationship between the
adjective and the entity ("is", "was", "tasted", etc.), as above. Using
this method, we collect 37 restaurant, 30 cuisine, 247 food, 45
service, and 56 staff patterns for positive and 18 restaurant, 9
cuisine, 221 food, 75 service, and 61 staff patterns for
negative. Table \ref{table:patts} shows example patterns in each class
for the food and staff entity types.

\begin{table}[h]
\begin{centering}
\begin{small}
\begin{subtable}[h]{1.0\linewidth}
\ra{1.3}
\begin{tabular}
{@{}p{3.75cm}|p{3.5cm}@{}}
\toprule
\bf Positive & \bf Negative \\ \hline
{ \sc insanely good \newline
simply perfect \newline
ridiculously good \newline
also incredible \newline
my fav \newline
perfectly crisp \newline 
definitely unique \newline 
always so fresh \newline 
just phenomenal \newline 
so decadent \newline 
highly addictive \newline
consistently great \newline 
wow amazing \newline 
perfect little \newline 
expertly prepared \newline 
freshly baked } &
{\sc almost raw \newline
very fatty  \newline
previously frozen \newline
comically bad \newline
absolutely awful \newline
not palatable \newline
fairly tasteless \newline
pretty generic \newline
so mediocre \newline
so bland \newline
still raw \newline
barely warm \newline
prepackaged frozen \newline
most pathetic \newline
sickly sweet \newline
luke warm} \\
\bottomrule        
 \end{tabular}
 \caption{Sample Learned Adjective Patterns for Foods}
   \label{table:food-patts}
   
\vspace{0.5cm}

\begin{tabular}[h]
{@{}p{3.75cm}|p{3.5cm}@{}}
\toprule
\bf Positive & \bf Negative \\ \hline
{\sc super helpful \newline
incredibly friendly \newline
super nice \newline
very personable \newline
so good \newline
so gracious \newline
very knowledgeable \newline
so kind \newline
extremely professional \newline
also fabulous \newline
even better \newline
still awesome \newline
always warm \newline
always attentive \newline
absolutely best \newline
our sweet 
} & 
{ \sc 
not apologetic \newline
not knowledgeable \newline
very rude \newline
too busy \newline
friendly enough \newline
just horrible \newline
not attentive \newline
very push \newline
more interested \newline
too lazy \newline
even worse \newline
every single \newline
very poor \newline
so few \newline
still no \newline
very unhappy
} \\
\bottomrule   
 \end{tabular}
  \caption{Sample Learned Adjective Patterns for Staff}
   \label{table:staff-patts}      
 \end{subtable}
 \end{small}
  \caption{Sample Learned Adjective Patterns}
 \label{table:patts}
 \end{centering}
\end{table}

\subsection{Creative Templates}

Finally, for our creative templates, we sample from our set of delexicalized 
sentences for each entity type, as long as they:
    \begin{itemize}  \setlength\itemsep{-0.5em}
\item contain a single AutoSlog adjective pattern
\item contain a single identifiable entity type
\item are between 5-15 words long
    \end{itemize}

We enforce these limitations to gather simple sentences that are short enough to
templatize. Thus, we end up with sentence templates for each entity type for both the positive
and negative classes, collecting 146 restaurant, 61 cuisine, 743 food, 90
service, and 144 staff patterns for positive and 45 restaurant, 12
cuisine, 480 food, 126 service, and 89 staff patterns for
negative. Table \ref{table:template-sentences} shows examples of our templatized
sentences for the positive and negative classes, with their AutoSlog-TS adjective patterns between brackets,
and capitalized subject extractions when applicable.  To construct a full review of a certain polarity, we
randomly select a sentence from the sets for each entity type.

\begin{table*}[ht]
\begin{footnotesize}
\begin{tabular}
{@{}p{1.6cm}|p{13.8cm}@{}}
\toprule
{\bf  Entity } & {\bf Template } \\ \midrule
    \multicolumn{2}{l}{ \bf Positive} \\ \midrule
{\sc restaurant}  &  {By  [FAR MY]  favorite  {\sc $<$RESTAURANT\_ENTITY$>$}  I HAVE EVER been to in my life .} \\ \hline
{\sc cuisine}  & { Wow what a great  [LITTLE  {\sc $<$CUISINE\_ENTITY$>$} ]  joint !} \\ \hline
{\sc food}  &  { The  {\sc $<$FOOD\_ENTITY$>$}  is not cheap , but  [WELL WORTH]  it.} \\ \hline
{\sc service}  & { The  {\sc $<$SERVICE\_ENTITY$>$}  is  [ALWAYS FRIENDLY]  and fast . } \\ \hline
{\sc staff}  & { {\sc $<$STAFF\_ENTITY$>$}  was  [EXTREMELY HELPFUL]  and knowledgeable and was on top of everything. } \\ \hline

    \multicolumn{2}{l}{ \bf Negative} \\ \midrule
{\sc restaurant}  &  { I was appalled by the experience and will  [NOT FREQUENT]  this  {\sc $<$RESTAURANT\_ENTITY$>$}  ever again.} \\ \hline
{\sc cuisine}  &  { [ITS YOUR]  typical  {\sc $<$CUISINE\_ENTITY$>$}  buffet , nothing to rave about .} \\ \hline
{\sc food}  &  { {\sc $<$FOOD\_ENTITY$>$}  smelled  [VERY BAD]  and tasted worse .} \\ \hline
{\sc service}  &  { We waited another 5 minutes ,  [STILL NO]   {\sc $<$SERVICE\_ENTITY$>$}  .} \\ \hline
{\sc staff}  &  { I went with 5 friends and our  {\sc $<$STAFF\_ENTITY$>$}  was  [REALLY RUDE]  .} \\
\bottomrule         
 \end{tabular}
 \caption{Examples of Learned Creative Sentence Templates by Entity and Polarity}
   \label{table:template-sentences}
 \end{footnotesize}
\end{table*}

We hypothesized that the creative templates would optimize stylistic
variation and hence interestingness, but that they would also include
cases that would require further refinement, or perhaps elimination by
a subsequent ranking phase. Since our focus here is on overgeneration,
we include these and evaluate their quality. Table \ref{table:templates} shows examples of each template type we
create. 

\begin{table}[h]
\begin{footnotesize}
\begin{tabular}
{@{}p{2cm}|p{5cm}@{}}
\toprule
{\sc Basic} & {The bar is beautiful. They had authentic japanese cuisine. The udon looked excellent. The hosts is dedicated. They had reliable customer service.}  \\ \midrule
{\sc Hyperbolic} & {The bar is also very fresh. They had delicious authentic japanese cuisine. The udon looked so delicious. The hosts is also very friendly. They had such amazing customer service.} \\ \midrule
{\sc Creative} & {This is by far my favorite bar in town. plus there is a great japanese cuisine grocery store that has tons of stuff. The udon is always fresh, delicious and made to order. Hosts was super friendly, looking forward to coming back and trying more items. The customer service is great and the employees are always super nice! } \\ \bottomrule                  
 \end{tabular}
 \caption{Examples of Instantiated Positive Review Variations}
   \label{table:templates}
 \end{footnotesize}
\end{table}

\section{Evaluating Template Styles}

In order to evaluate our template variations, we choose to focus on three particular criteria: {\it convincingness,
interestingness,} and {\it naturalness}. We evaluate {\it convincingness} because creative language such as hyperbole
is often used in persuasive language, along with other figurative forms \cite{KreuzRoberts95}. {\it Naturalness} is an important concern in generation, so we are also interested in the comparison between the perceived naturalness of each variation style, and we hypothesize that {\it interestingness} would increase as we used more content from organic reviews in our {\sc hyperbolic} and {\sc creative} templates.

To create an evaluation dataset, we instantiate each template type with entities from a hypothetical MR
in one of seven popular cuisine types to standardize the content, as illustrated in Table \ref{table:templates}. For example, sample slot values could be: {\sc \{restaurant[bar],
cuisine[Japanese], food[udon], service[customer service],
staff[hosts]\}}.

Our objective is to evaluate whether we can improve upon vanilla-style
hand-crafted templates for restaurant reviews by utilizing in
hyperbolic and creative elements of organic reviews that we
harvest. We set up an annotation experiment on Amazon Mechanical
Turk\footnote{\url{https://www.mturk.com/}}, where each Human
Intelligence Task (HIT) presents Turkers with a sample of our three
review variations, all of the same polarity and instantiated with the
same entities. Turkers are asked to judge the reviews based on three
criteria: {\it convincingness} (Do you believe the opinion given?),
{\it interestingness} (Is the review engaging?), and {\it naturalness}
(Is the review coherent?). Turkers are asked to rate each review on a
three point scale ({\it high, medium} and {\it low}) for each
criteria. We release 200 variation triples (100 per polarity class)
and ask for five judgements per HIT, tagging a review with a quality if
the majority of annotators agree on it (i.e. 3 or more
Turkers). Average agreement for individual Turkers with the majority
is above 73\%.

\begin{figure}[h] 
\begin{center}
   \includegraphics[width=\textwidth]{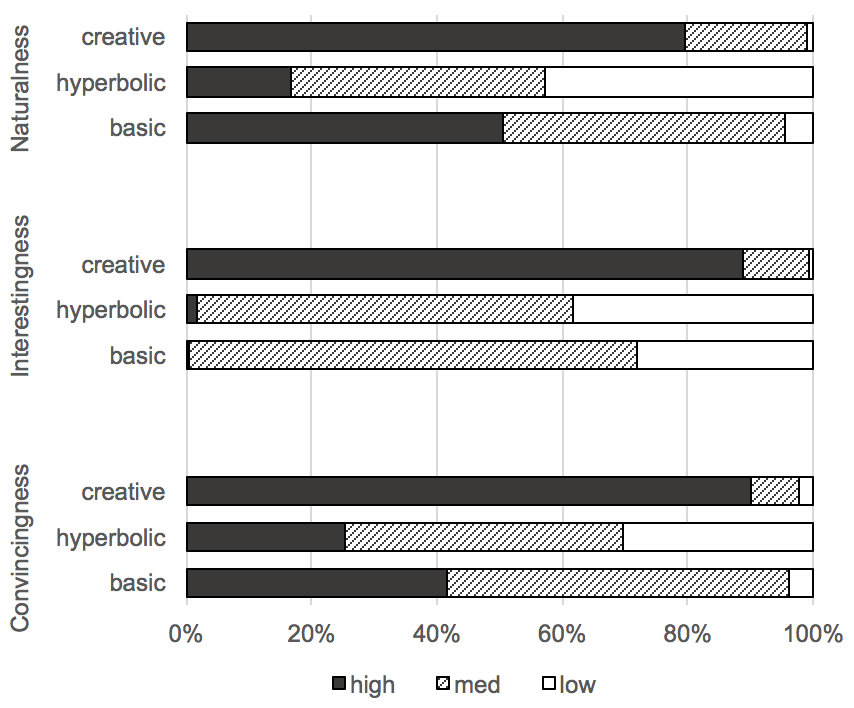}
  \caption{Distribution of Template Variations by Evaluation Criteria}
  \label{fig:evaluation}
  \end{center}
\end{figure}

Figure \ref{fig:evaluation} shows the distribution of {\it high,
  medium,} and {\it low} scores for each of the variation types for each
  criterion. From the results, we observe that for all
criteria, the {\sc creative} class has the highest distribution of
{\it high} majority votes. Interestingly, although we hypothesized
that the {\sc hyperbolic} reviews would be better received than the
{\sc basic} reviews, we observe that in fact the {\sc basic} reviews
receive more {\it high} votes on {\it convincingness}. We note that 
for the future ranking, more context information is necessary when
selecting appropriate hyperbolic patterns with which to modify the
{\sc basic} reviews. For example, if a learned pattern is {\sc other amazing},
the pattern should only be used when a set of items are being described, and not
stand-alone. Similarly, the {\sc basic} reviews are also more {\it natural} than the {\sc
  hyperbolic} ones, although both variation types score very similar
percentages for {\it medium} scores.

For the creative reviews, a crucial next step for ranking is to consider context and develop heuristics for 
finding the most appropriate entities for lexicalization. For example, for very specific creative templates such as: {\it "I also got one that HAD NOT been separated , so it was  [JUST HALF]  of a  {\sc $<$FOOD\_ENTITY$>$ }."}, or {\it "The {\sc $<$FOOD\_ENTITY$>$} were similarly a mix of nearly raw to overly crisp."}, it is necessary to select food items similar to the original instantiations, or to characterize and classify entities based on specific properties.

Given the high appeal of the {\sc creative} reviews on all counts, we are interested in more closely exploring examples in the data. Table \ref{table:creative-examples} shows two examples of {\sc creative} reviews: one that received {\it high} scores on all criteria, and one that received majority (no creative review received all {\it lows}). It is clear that the biggest disconnect in the low-scoring {\it creative} review is the coherence between sentences, which as an important next step to consider as future work given the proof-of-concept presented here. We also note that we can also improve the fully high-scoring review by fixing grammatical errors and applying more informed content selection.

\begin{table}[t]
\begin{footnotesize}
\begin{tabular}
{@{}p{1cm}|p{6cm}@{}}
\toprule
{\sc All \newline High} & {It is one of my favorite cafe in las vegas. Thank you irma for your amazing mediterranean cuisine cooking! I am always amazed at how fast my falafel arrives. Victor the owner was super nice and cordial our hosts norma was also. Always a great place to go and service is always amazing!} \\ \midrule
{\sc Mostly \newline Low} & {It's just too bad that the bar itself is not better. Very bad american cuisine.... Guess what came on top of my hotdog? I took my family there for father's day and the hosts was so rude. 555 pm still no customer service.} \\ \bottomrule                  
 \end{tabular}
 \caption{Example of {\it High} and {\it Low} Rated Creative Reviews}
   \label{table:creative-examples}
 \end{footnotesize}
\end{table}

To get a better sense of how grammatically correct the review template variations are, we conduct another evaluation study where we present Turkers with the same set of reviews, and ask them to rate each review based on the content (checking subject-verb agreement, plurality, tense, etc.). Similar to the previous study, we gather 5 judgements for each set of three variations, and aggregate results using majority vote. Average agreement for individual Turkers with the majority in this task is above 80\%, higher than the more subjective study on convincingness, interestingness, and naturalness.

 \begin{figure}[b] 
\begin{center}
   \includegraphics[width=\textwidth]{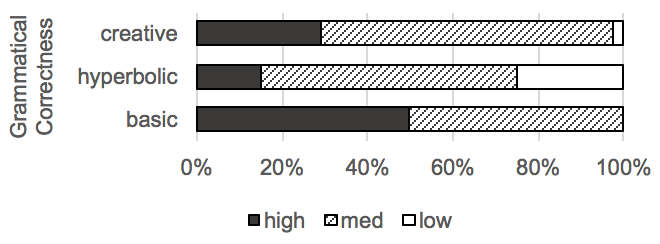}
  \caption{Distribution of Votes on Template Variation Grammar}
  \label{fig:grammar-eval}
  \end{center}
\end{figure}

Figure \ref{fig:grammar-eval} shows the results of the study. We find that for all three variations, the {\it med} class receives the majority of the votes, but that the {\sc basic} reviews are the most grammatically correct (since the templates are designed, not harvested). Similarly, the {\sc hyperbolic} reviews have the largest percentage of {\it low} scores, since their creation involves modifying templates with learned adjectives. Ranking the best patterns/sentences to use will allow us to improve the grammatical coherence of the templatized utterances for the {\sc hyperbolic} and {\sc creative} classes.

To better understand the linguistic characteristics of the creative reviews by class, we run the Linguistic Inquiry and Word Count (LIWC) tool \cite{Pennebaker15} on the full set of 100 {\sc positive} and 100 {\sc negative} {\it creative} reviews. When comparing the linguistic categories for each class, we find that the difference between the {\sc positive} and {\sc negative} reviews are significant ($p < 0.05$, t-test) for many of the categories. Table \ref{table:liwc} shows some of the most interesting categories\footnote{All of the categories are statistically significant, and are shown in order of most to least significant.}.

\begin{table}[t]
\begin{small}
\begin{tabular}
{@{}p{3.3cm}|p{3.3cm}@{}}
\toprule
\bf Positive & \bf Negative \\ \hline
{\sc Affective Proc. \newline
Exclamations \newline
Friends \newline
$1^{st}$ Person Singular \newline
Achieve \newline
Certainty \newline
Biological Proc. \newline
Ingestions \newline
Insight \newline
Reward 
} & 
{ \sc Differentiation \newline
Risk \newline
$1^{st}$ Person Plural \newline
Anxiety \newline
Adverbs \newline
Anger \newline
Social Proc. \newline
$2^{nd}$ Person \newline
Motion \newline
Cognitive Proc.  } \\ \bottomrule         
 \end{tabular}
 \caption{Statistically Significant LIWC Categories by Polarity}
   \label{table:liwc}
 \end{small}
\end{table}

On average, the {\sc positive} templates are characterized by word classes that exemplify achievement (e.g. {\it "even better", "champion"}) and certainty (e.g. {\it "always excellent", "absolutely amazing",} and {\it "definitely my go-to place"}).
As well as $1^{st}$ person statements relating to use of
the senses (affective processes like {\it "my favorite place to get rice in Las Vegas!"}, biological processes ({\it "I just had the most amazingly delicious and freshly prepared couscous!"}), and
ingestion ({\it "good, tasty comfort pizza"}). 

The negative contains more oppositional language directed
at the second person, often as advice ({\it "you can get a much better pizza elsewhere at far less cost."}), with categories like differentiation ({\it "but it's not great"}), and strong
emotion indicators like anxiety ({\it "horrible service, finally just left"}) and anger ({\it "I was so angry that I contacted the restaurant manager"}).

\section{Conclusions}

In this paper, we show that we can construct convincing, interesting,
and natural restaurant review templates by using a data-driven method
to harvest highly descriptive sentences from hyperbolic restaurant
reviews. We generate three variations of review templates, ranging
from very basic, to hyperbolic, to very creative, and show that
the creative ones are more appealing to readers than the others. Future work 
will focus on ranking the candidate sentence templates we harvest to 
improve review coherence. As we develop better templates,
we will evaluate them against baselines from existing NLG systems to guide our 
generation of more exciting and expressive stylistically varied reviews.

\bibliography{nl,phd,emnlp2017}
\bibliographystyle{emnlp_natbib}

\end{document}